# Flow Reconstruction from Sparse Measurements in Urban Drainage Networks: A Simulation-Based Evaluation of Data-Driven Sparse Sensing

Zihang Ding[a], Amit Kumar[a], Imran Md. Azizul Islam[a], Mila Avellar Montezuma[b], Ruihang Zhang[c], Kun Zhang[a*]

[a]Department of Civil and Environmental Engineering, University of Minnesota Duluth, Duluth, Minnesota, United States

[b]IHE Delft Institute for Water Education, Delft, Netherlands; Rede CLIMA-Brazilian Research Network on Global Climate Change (Ministry of Science Technology and Innovation of Brazil), Brazil

[c]Department of Mechanical and Industrial Engineering, University of Minnesota Duluth, Duluth, Minnesota, United States

*Corresponding author: kunzhang@d.umn.edu, ORCID: https://orcid.org/0000-0002-1062-8323

**Abstract**: Urbanization and increasingly frequent intense storms are placing stress on urban drainage networks. While dense monitoring is desirable, practical constraints in time, budget, and technology hinder its full implementation. How to monitor and predict flow conditions across the entire network under constrained resources is a major challenge. To address this, we evaluated data-driven sparse sensing (DSS), which selects sensor locations and reconstructs network-wide flow from the same reduced basis, in a 76-conduit urban drainage network. Simulations of a calibrated SWMM model were used to construct an event-weighted spatial basis using singular value decomposition (SVD), select rank-specific layouts using pivoted QR, and define the reconstruction decoder. Applied to 225 held-out simulations combining 25 plausible calibrated parameter sets with nine rainfall events, monitoring 5 of the 76 conduits (6.6%) achieved a median system-level Nash-Sutcliffe efficiency (NSE) of 0.992 ($10^{th}$ percentile 0.973) for observed storms, and a median NSE of 0.967 at the active unmonitored conduits. Pivoted QR performed comparably to genetic-algorithm layouts without requiring iterative layout searches, whereas largest-upstream-area placement failed with the DSS decoder and trailed QR even with a weighted regression decoder. The DSS flow reconstruction performance was insensitive to prescribed multiplicative Gaussian measurement noise, additive noise up to about 2 L/s and a 10 L/s detection floor, but showed location-dependent responses to loss of a monitored conduit, which a refitted regression decoder largely removed. This study supports the use of DSS, driven by physics-based simulations, for deciding how many flow sensors to install and where.

## Highlights

- Data-driven sparse sensing (DSS) identified conduits to monitor storm sewers
- Monitoring 5 of 76 conduits effectively reconstructed network-wide flow rates.
- DSS matched search-based placement and outperformed largest-upstream-area placement
- DSS's reconstruction performance was insensitive to moderate measurement noise
- A refitted regression decoder mitigated the loss after sensor failure

## 1. Introduction

Urban drainage systems are increasingly exposed to compound pressures from climate change, rapid urbanization, and aging infrastructure, which have substantially increased the frequency and severity of surcharge, backwater effects, and pluvial flooding in many cities (Berndtsson et al., 2019; Fletcher et al., 2024). Effective mitigation and real-time operation require accurate monitoring and prediction of distributed hydraulic states across sewer networks, such as water depths, flow rates, and overflow-related conditions (Fu et al., 2022; García et al., 2015). However, dense sensor deployment across large-scale urban drainage networks remains economically and operationally difficult because hydraulic conditions change rapidly across connected junctions and conduits, while field instrumentation is constrained by cost, accessibility, installation and maintenance requirements and sensor reliability (Pleau et al., 2022; Pritsis et al., 2026; Sebicho et al., 2024; Tao et al., 2024). Two practical questions remain: 1) where are the best locations to take measurements within the drainage network? And 2) how much of the network flow state can be reconstructed from measurements at a subset of locations?

The first question – sensor placement – has been approached using network observability, clustering, information-theoretic or value-of-information criteria, and optimization-based experimental design across sewer, drainage, flood, and related water networks (Alfonso and Price, 2012; Banik et al., 2017a, 2017b; Crowley et al., 2025; Farahmand et al., 2022; Huang et al., 2025; Krause et al., 2008; Li, 2021; Sambito et al., 2020; Wang et al., 2023; Zheng et al., 2025). Some flood-monitoring studies have also optimized sensor locations for calibration or early warning under explicit cost and parsimony objectives (Fattoruso et al., 2015; Grimaldi et al., 2024).

The second question – flow reconstruction and prediction at unmonitored locations under sparse monitoring – has also been approached using methods such as graph-based deep-learning models (He et al., 2025a, 2025b; Zheng et al., 2026) or topology-driven placement coupled with graph-based hydraulic-state reconstruction (He et al., 2026). Related virtual-sensing and reduced-order approaches in other dynamical systems likewise reconstruct distributed states from limited measurements when the underlying representation is informative (Ercan et al., 2023; Luo et al., 2023; Wang et al., 2024). Together, these studies show that sparse monitoring can support network-wide state estimation, although useful monitoring locations depend on the hydraulic variable and monitoring task being considered (Pritsis et al., 2026).

Data-driven sparse sensing (DSS) provides a direct way to connect sensor placement with network-wide flow reconstruction. DSS uses singular value decomposition (SVD) to identify dominant spatial patterns in a development dataset and pivoted QR factorization to select locations that efficiently sample those patterns (Manohar et al., 2018). Measurements from the selected locations are then used to estimate the retained SVD coefficients and reconstruct the modeled network state, so sensor placement and reconstruction are based on the same reduced representation. DSS has been applied to groundwater-level networks, regional streamflow, and nutrient monitoring, where low-dimensional hydrologic structure has been used to support reconstruction from limited observations (Bin Mamoon et al., 2025; Ohmer et al., 2022; Zhang et al., 2023b). More broadly, compressed sensing provides a distinct sparse-recovery formulation (Donoho, 2006) that has also been used to reconstruct hydrologic streamflow and concentration time series from limited observations (Zhang et al., 2023a). To our knowledge, DSS has not been tested in urban drainage, where rainfall-runoff generation and routing through connected pipes produce rapid, event-driven, and spatially heterogeneous flow responses (He et al., 2025a, 2025b). Whether a reduced representation learned from a limited development catalog remains effective under unseen hydrologic conditions is therefore an important question for urban drainage applications.

Reconstruction under intact monitoring conditions is only part of the monitoring problem. Practical flow monitoring is subject to various uncertainties such as measurement noise and sensor failures. These disturbances can impact flow reconstruction differently, particularly given that individual monitoring locations can differ in both their information contribution and redundancy within a sensor network (Crowley et al., 2025; Zheng et al., 2025). Post-deployment analyses further indicate that the impact of individual sensors on network-wide monitoring may vary (Zheng et al., 2026). Consistent with this, research on sensor failures has shown that failures of high-importance sensors can significantly degrade monitoring performance (Zhang et al., 2020), while redundant measurements have been used to support sensor-fault detection in urban drainage systems (Pleau et al., 2022). In addition to evaluating the robustness of flow reconstruction to these uncertainties, identifying sensor configurations that are sensitive to measurement uncertainties is also critical as it informs better sensor placement.

In this study, we tested DSS in an EPA-SWMM drainage network with 76 conduits in Duluth, Minnesota. The primary objective was to determine whether and how effective a spatial basis, sparse layout, and reconstruction decoder constructed from simulations generated with a calibrated SWMM parameterization could reconstruct modeled conduit flow when applied to held-out rainfall events and parameter sets. We first calibrated the SWMM parameterization using observed storm events. With another 17 storm events as input, the calibrated model was then used to construct an SVD-derived spatial basis and rank-specific QR layouts. We then evaluated this DSS model on 225 simulations combining 25 plausible calibrated parameter sets with 9 rainfall events, using the spatially sparse measurements obtained from the subset of conduits to reconstruct system-wide flow rates. Additional analyses compared QR with heuristic and search-based placement methods and the DSS decoder with regression decoders, and evaluated robustness to measurement errors and to the loss of a monitored conduit.

## 2. Data and Methods

### 2.1. Data-Driven Sparse Sensing

DSS integrates SVD and QR factorization to identify informative sampling locations and reconstruct distributed system states. In brief, SVD can identify a low-dimensional space onto which the signal (e.g., conduit flow states in the storm sewer system, organized as a matrix) is projected, while QR factorization with column pivoting selects sensor locations that efficiently sample the retained basis.

Natural signals, including both 1-D temporal data (e.g., runoff hydrographs) and 2-D spatially distributed data (e.g., spatial maps of flow rates), can be represented by discrete state vectors $\boldsymbol{x_i}$. Each system state $\boldsymbol{x_i} \in \mathbb{R}^n$ can be represented by a linear combination of appropriate basis vectors arranged into a matrix $\boldsymbol{\Psi_r}$, with amplitudes, $\boldsymbol{a_i} \in \mathbb{R}^r$, i.e.,

$$\boldsymbol{x_i} = \boldsymbol{\Psi_r a_i} \tag{1}$$

where $i$ indexes the system states or snapshots, $n$ is the dimension of the full state, $r$ is the number of retained basis vectors, $\boldsymbol{\Psi}_r \in \mathbb{R}^{n\times r}$ contains the retained basis vectors, and $\boldsymbol{a}_i$ contains the corresponding coefficients for state $\boldsymbol{x}_i$.

Many natural signals can be represented compactly when an appropriate basis is used. Often, a generic or universal basis, such as Fourier or wavelets, can provide compact representations without prior knowledge of the signal. However, these bases are not tailored to the spatio-temporal structure of the system and may therefore require more retained components or measurements for a given reconstruction accuracy. When representative observations or computational data are available, a basis can instead be derived directly from the system states.

In this study, we developed an SVD-derived spatial basis tailored to the development data, which provides the best rank-r least-squares approximation of those data (Figure 1). That is, given a training dataset containing computational states arranged in the matrix $\boldsymbol{X}$, the SVD

$$\boldsymbol{X} = \boldsymbol{\Psi\Sigma V^T} \tag{2}$$

identifies the orthonormal spatial basis $\boldsymbol{\Psi}$, while $\boldsymbol{\Sigma}$ holds the singular values, and $\boldsymbol{V^T}$ comprises the corresponding temporal coefficients. Hereafter, this SVD-derived spatial basis is referred to as the DSS basis. The first $r$ vectors in $\boldsymbol{\Psi}$, i.e., $\boldsymbol{\Psi}_r = [\psi_1, \psi_2, \dots, \psi_r]$ represent the retained spatial modes for the dataset. Given a limited set of measurements $\boldsymbol{y}$ sub-sampled from the target network state $\boldsymbol{x}$, where $\boldsymbol{y} = \boldsymbol{C}\boldsymbol{x}$ and $\boldsymbol{C}$ is a sampling operator representing the locations at which measurements are taken, the coefficient vector $\widehat{\boldsymbol{a}}$ can be estimated from measurements as

$$\widehat{\boldsymbol{a}} = (\boldsymbol{C}\boldsymbol{\Psi}_{\boldsymbol{r}})^{\dagger}\boldsymbol{y} \tag{3}$$

to yield the estimate for the target network state

$$\widehat{\boldsymbol{x}} = \boldsymbol{\Psi}_{\boldsymbol{r}}\widehat{\boldsymbol{a}} \tag{4}$$

where $(\cdot)^{\dagger}$ denotes the Moore–Penrose pseudoinverse and $\widehat{\boldsymbol{x}}$ denotes the reconstructed state.

Importantly, the locations where measurements are taken can be selected to efficiently sample the retained basis modes in $\boldsymbol{\Psi}_r$. These sampling locations can be obtained using QR factorization with column pivoting (Manohar et al., 2018),

$$\boldsymbol{\Psi}_{\boldsymbol{r}}{}^{\boldsymbol{T}}\boldsymbol{\Pi} = \boldsymbol{Q}\boldsymbol{R} \tag{5}$$

where $\boldsymbol{\Pi}$ is the column-permutation matrix produced by QR pivoting. The first $m$ pivot indices define the selected sampling locations and were used to construct the sampling operator $\boldsymbol{C}$.

In this study, the SVD-derived spatial basis ($\boldsymbol{\Psi}_{\boldsymbol{r}}$) is used to represent modeled network states and pivoted QR factorization is used to identify informative sampling locations, which define the sampling operator $\boldsymbol{C}$. Measurements collected at these locations in the stormwater network are then used with the retained basis to reconstruct the network-wide modeled conduit flow states.

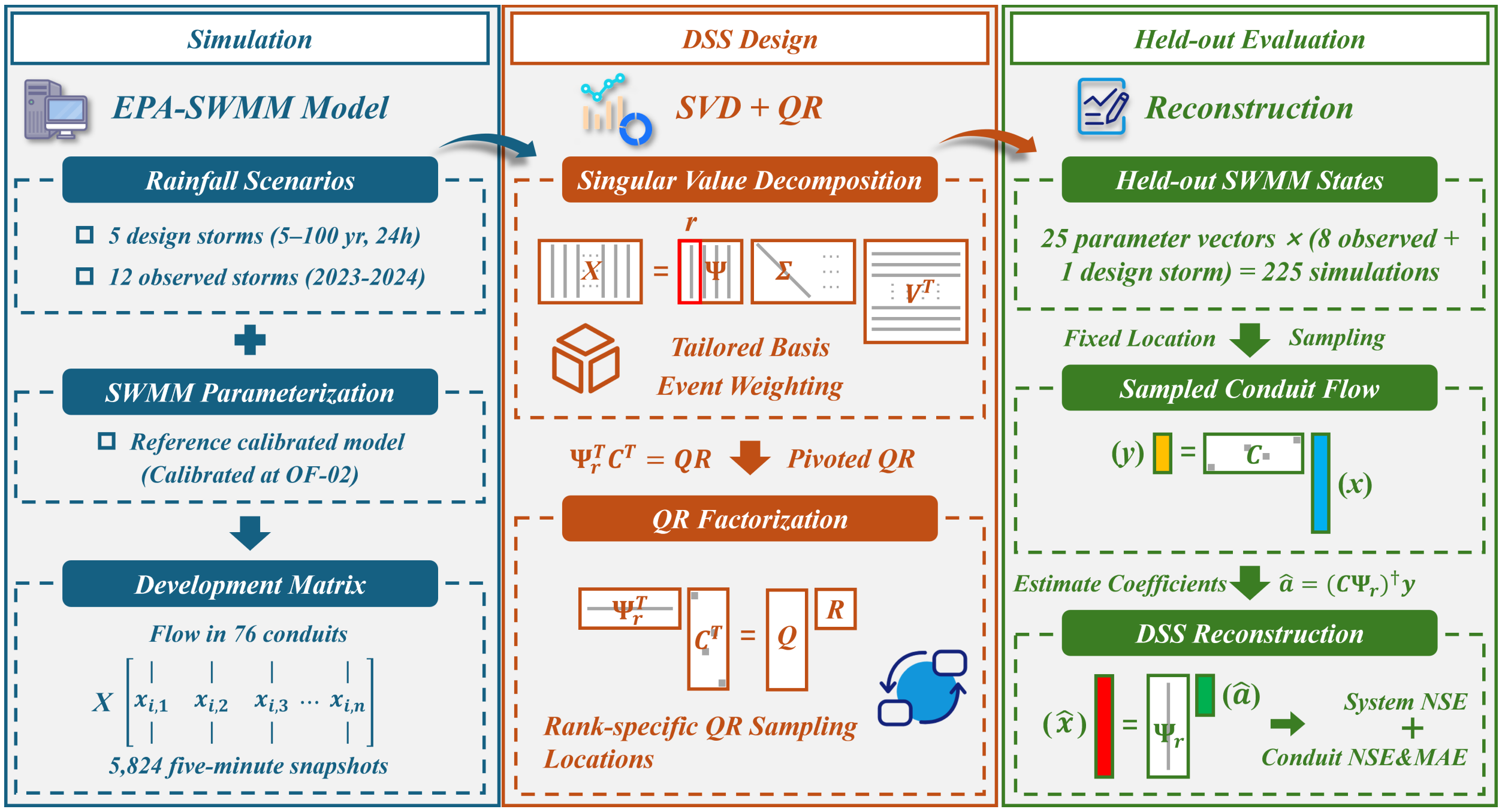

**Figure 1.** Workflow of the data-driven sparse sensing (DSS) framework for modeled conduit-flow reconstruction in 76 conduits: simulation of 17 development storms with the calibrated SWMM model, DSS development (event-energy-weighted SVD basis, rank-specific QR sampling layouts and linear decoder) and held-out evaluation on 225 simulations (25 parameter sets × 9 storms).

## 2.2. Case Study

This study focuses on the Woodland catchment in Duluth, Minnesota, which covers 133.53 ha (1.34 km²) and is characterized by steep terrain and a connected storm-sewer network (Figure 2). Runoff from 48 subcatchments is routed through 77 nodes (71 junctions, 4 outfalls and 2 storage units), 70 pipes 0.30–1.37 m in diameter (3.07 km in total), 6 open channels (0.56 km) and 54 street links (3.30 km) (Figure S1). The 76 pipes and open channels served as candidate sampling locations and form the reconstructed state; the street links (surface flow paths) are neither monitored nor reconstructed.

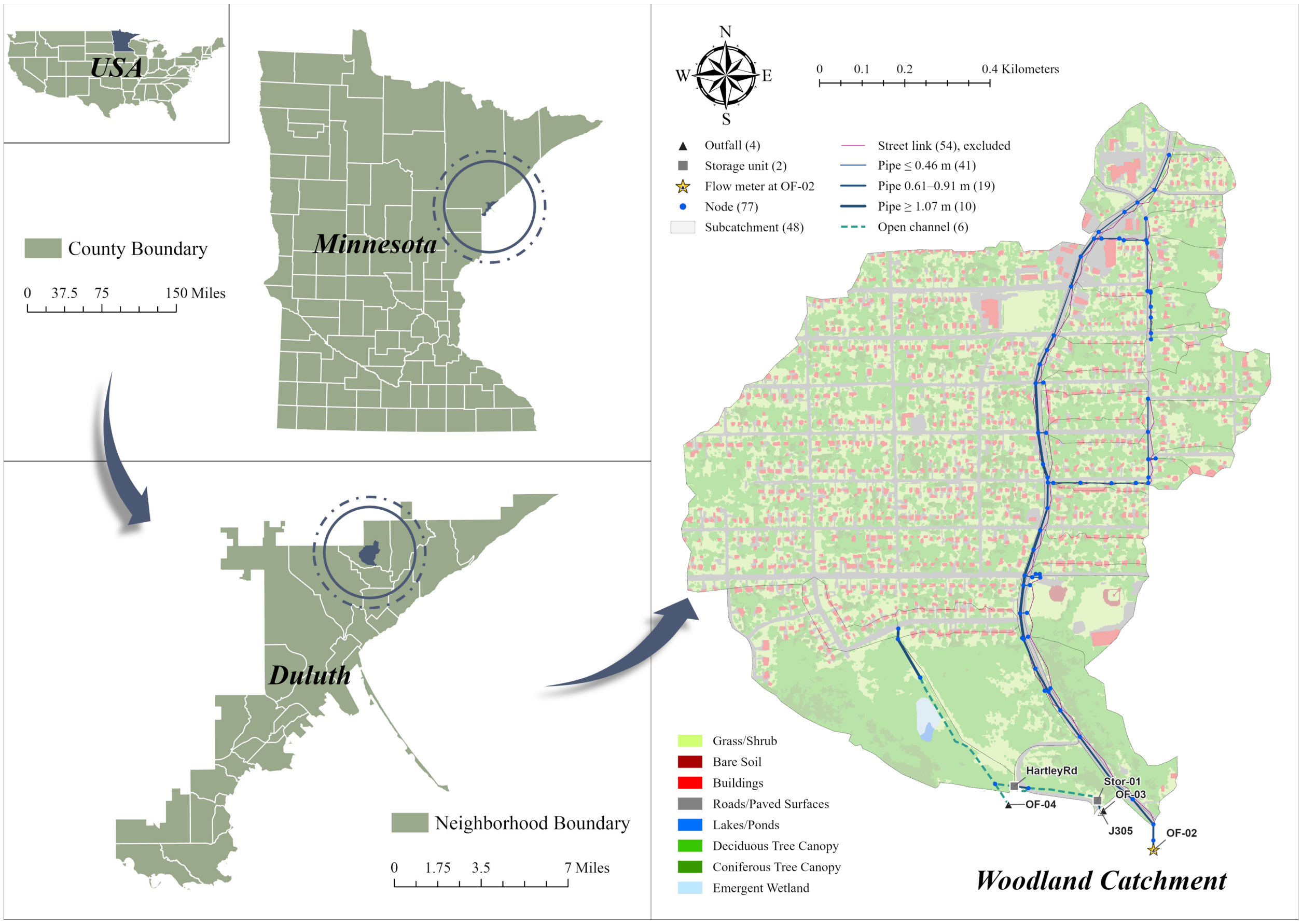

**Figure 2.** Location, land cover, subcatchments and drainage network of the Woodland catchment, Duluth, Minnesota, USA.

Imperviousness was calculated for all 48 subcatchments using pixel-polygon area weighting. A 2015 1-m classified land-cover raster was used to identify buildings and paved surfaces as impervious, and the baseline imperviousness was updated using the 2015–2024 change in 30-m fractional imperviousness from the USGS Annual NLCD Collection 1 Science Products (U.S. Geological Survey (USGS), 2024). The resulting area-weighted catchment imperviousness increased from 25.260% to 26.247%.

Subcatchment slope was calculated from the Second Generation Seamless LiDAR bare-earth elevation data provided by the Minnesota Geospatial Information Office (n.d.). The original 0.5-m elevation raster was aggregated to 2 m, and percent-rise slope was calculated using the Horn 3 × 3 method (Horn, 1981). Subcatchment mean slopes ranged from 1.642% to 21.322%, with an area-weighted catchment mean of 8.624%.

## 2.3. SWMM Model Calibration and Validation

We used EPA-SWMM 5.2 (Rossman and Simon, 2022) to generate the conduit-flow fields across the network used for DSS development and evaluation. Rainfall was assumed to be uniform across the catchment, from a single rain gauge, consistent with the spatial rainfall resolution of about 2 km recommended for urban catchments of about 100 ha (Berne et al., 2004).

The SWMM model was calibrated using discharge measurements at OF-02, the only outfall of the catchment with a flow meter, and validated against a separate storm event at the same location. At OF-02, an area-velocity meter continuously measured the water depth and velocity at 5-min intervals from June 8 to June 18, 2024. Flow was calculated from synchronized depth and velocity measurements using the wetted cross-sectional area and was converted to event-excess flow after baseline separation. The meter recorded three complete storm events: June 11 (11.33 mm), June 12 (6.34 mm), and June 15–16 (7.27 mm). Events on June 11 and June 15-16 were used to calibrate the model and that on June 12 was used for validation.

Twenty-one parameters related to infiltration and runoff routing were calibrated, including zoned values of subcatchment width, impervious and pervious roughness, depression storage, impervious-to-pervious routing, and Horton infiltration. The subcatchment impervious fractions and slopes were extracted from the GIS datasets mentioned above, and the parameters related to sewer hydraulics were estimated based on site surveys and information extracted from the sewer system database. The NSGA-II evolutionary algorithm (Deb et al., 2002) was used to calibrate the hydrologic parameters using five fixed random seeds and 3,000 SWMM evaluations per seed, giving 15,000 evaluations in total. The reference calibrated parameterization was selected from the Pareto front by equal-weight normalized distance to an ideal point using 3 criteria (calibration-event NSE, a composite hydrograph score, and peak and volume errors). The parameter zoning, calibration bounds, and selection procedures are provided in Supplementary Information Text S1 and Tables S1–S2.

The selected reference calibrated parameterization yielded NSE = 0.559 and 0.553 for calibration events and NSE = 0.612 for the validation event, with peak-flow and event-volume biases of −46.1% and −42.1% (Figure 3).

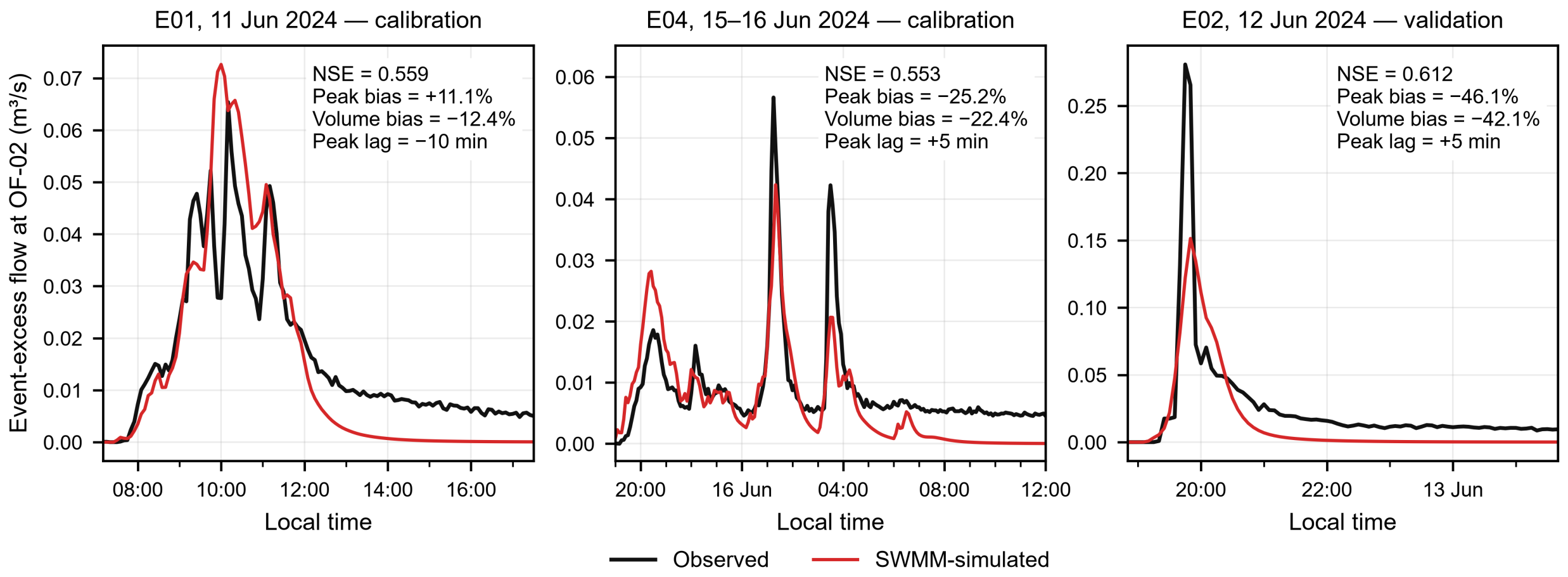

Figure 3. Reference SWMM model at the OF-02 flow meter: observed and simulated event-excess flow for the calibration events E01 (June 11, 2024) and E04 (June 15–16, 2024) and the validation event E02 (June 12, 2024), at 5-min steps over the score window (local time). Event-excess flow is flow above the pre-event baseline.

## 2.4. DSS Model Development

SWMM simulations from the selected reference parameterization were used to develop the DSS basis: 5,824 five-minute snapshots of the flow in the 76 conduits for 17 storms, including 12 observed storms from 2023–2024 and five 24-h design storms with return periods of 5–100 years; each storm was simulated from 6 hours before rainfall onset to 6 hours after rainfall end. The snapshots were concatenated without mean centering, after the snapshot block of each storm was scaled to unit Frobenius norm, also called event-energy weighting (or event-weighted basis construction). Event-energy weighting was preferred over snapshot-equal weighting because it ensures each storm contributed equally to the SVD. Under snapshot-equal weighting the five design storms carry 92.0% of the squared development flow in 37.1% of the snapshots. Event-energy weighting also outperformed snapshot-equal weighting in a cross-validation on the development storms (13 folds; mean validation NSE 0.954 versus 0.842; Text S6 and Figure S5).

After that, SVD was performed on this dataset to construct the DSS basis (Eq. 2), and pivoted QR factorization was further applied to identify informative sampling locations (Eq. 5) for retained rank $r$ = 1–10. We set the number of monitored conduits equal to the retained rank ($m = r$) to construct a rank-matched square sampling system. In addition, QR was applied independently at each retained rank $r$, so the resulting layouts were rank specific rather than nested.

## 2.5. Held-out Reconstruction and Supporting Analyses

Held-out reconstruction was then performed to evaluate the method's ability to reconstruct flow rates across the network. This evaluation combined 9 rainfall events, including 8 observed storms and one 200-year 24-h design storm, with 25 plausible calibrated parameter sets obtained from NSGA-II during calibration, giving 225 SWMM simulations. A summary of the rainfall events used in this study is provided in Text S2 and Tables S3–S4.

DSS-based flow reconstruction was evaluated on the 225 held-out simulations using the rank-specific QR-selected monitoring layouts for $r$ = 1–10. Among different $m$ scenarios, $m$ = 8 (about 10% of the conduits) was used as an illustrative budget, and its results were further analyzed. Reconstruction performance was quantified using Nash–Sutcliffe efficiency (NSE; Nash and Sutcliffe, 1970):

$$\boldsymbol{NSE = 1 - \frac{\sum_{i=1}^{N}(x_i - \hat{x}_i)^2}{\sum_{i=1}^{N}(x_i - \bar{x})^2}} \tag{6}$$

where $x_i$ and $\hat{x}_i$ are the SWMM-simulated and DSS-reconstructed conduit flows, respectively, $\bar{x}$ is the mean of the SWMM-simulated values used in the corresponding NSE calculation, and $N$ is the number of values in that calculation. First, to evaluate performance across the entire network, "system-level" NSE was computed from the combined network–time state. For "system-level" NSE, index $i$ covers all conduit–time states from the 76 conduits and all available time steps within one held-out simulation. Second, to evaluate each conduit individually, "conduit-level" NSE was calculated separately for each conduit time series, and reported only for active scenarios, with a temporal standard deviation ≥ 10 L/s; mean absolute error (MAE) was reported for all conduits.

In addition, supporting analyses were further performed to compare pivoted QR with alternative placement and decoding methods, and to evaluate its sensitivity to measurement errors and to the loss of one monitored conduit:

1. **Reference placement and decoding benchmark**: We compared the DSS's flow rate reconstruction performance across monitoring layouts obtained from pivoted QR, Uniform Random, genetic algorithm (GA), and a largest-upstream-area heuristic for $m$ = 1–10. Each method determined monitoring locations using the development dataset and was evaluated using the same reconstruction procedure (using a DSS decoder). For GA-based reconstruction, the layout with the best development objective among 30 GA runs (GA-best) was used for each $m$. For Uniform Random placement, the corresponding median was calculated across all 76 one-conduit layouts at $m$ = 1 and 2,048 unique monitoring layouts for each $m$ = 2–10. GA minimized development-set reconstruction loss, and the largest-upstream-area heuristic monitored the $m$ conduits with the largest upstream drainage area. The DSS decoder was also compared with ridge-regression decoders (Hoerl and Kennard, 1970), a regularized alternative to the pseudoinverse in sparse sensing (Erichson et al., 2020; Jayaraman and Al Mamun, 2020; Karnik et al., 2026), unweighted and event-weighted, with $\lambda$ selected by the same storm-wise cross-validation (details in Text S4).

2. **Measurement-error sensitivity**: To evaluate the robustness of our reconstruction framework under noisy conditions, we applied graduated multiplicative Gaussian noise, drawn independently for each monitored conduit and time step, to the selected measurements as:

$$y_{noise} = y(1 + \sigma z) \tag{7}$$

$$z \sim \mathcal{N}(0,1), \quad \sigma = \frac{\eta}{3}, \quad \eta \in \{0.05, 0.10, 0.15\} \tag{8}$$

where $z$ is a standard normal random variable, $\eta$ is the prescribed noise level (5%, 10% or 15%), and the standard deviation is one-third of this level, so that about 99.7% of the perturbations lie within ±5%, ±10% or ±15% (three-sigma rule). Additive Gaussian noise with $\sigma$ = 0.5 – 10 L/s was also tested, together with two systematic errors: a detection floor of 10 L/s, below which measurements were set to zero, and a ±20% bias (measurements multiplied by 0.8 or 1.2) on all sensors or on one sensor at a time. For each sensor count ($m$ = 1–10), 1,000 Monte Carlo realizations were generated for each random-noise case (Text S5).

3. **Monitored conduit loss:** To understand how failures affect the reconstruction performance, we removed measurements from one monitored conduit at a time from each rank-specific DSS layout for $m$ = 2–10 and evaluated reconstruction using the remaining measurements. In the uncompensated case, the original rank-$r$ DSS basis was retained and the remaining $r - 1$ measurements were used to estimate the original $r$ coefficients through the Moore–Penrose pseudoinverse. In the truncated-basis case, the basis was truncated to the leading $r - 1$ modes to match the remaining measurements, and in the refitted-ridge case, an event-weighted ridge decoder refitted on the development data without the failed conduit was used (Text S5). The signed loss for sensor $i$, rank $r$, and held-out scenario $e$ was calculated as:

$$L_{i,r,e} = NSE_{r,e}^{baseline} - NSE_{i,r,e}^{loss} \quad (9)$$

where the baseline is the system-level NSE of the complete layout with the DSS decoder, and positive values indicate a decrease in reconstruction performance after measurement loss.

# 3. Results

## 3.1. System-Level Reconstruction Accuracy

System-level flow reconstruction performance generally improved as the number of monitored conduits $m$ increased (Figure 4a). For the 8 observed rainfall events, median NSE increased from 0.855 at $m$ = 1 to 0.934 at $m$ = 3, 0.992 at $m$ = 5 and 0.996 at $m$ = 8, while the corresponding medians for the 200-year design event were 0.598, 0.961, 0.977 and 0.988. The variability of reconstruction performance also decreased with $m$; the medians of the 8 observed events ranged from 0.877 to 0.964 at $m$ = 3 and from 0.972 to 0.997 at $m$ = 5. The density plots of simulated versus reconstructed flow showed the same progression, with values concentrating along the 1:1 line as $m$ increased (Figure 4b).

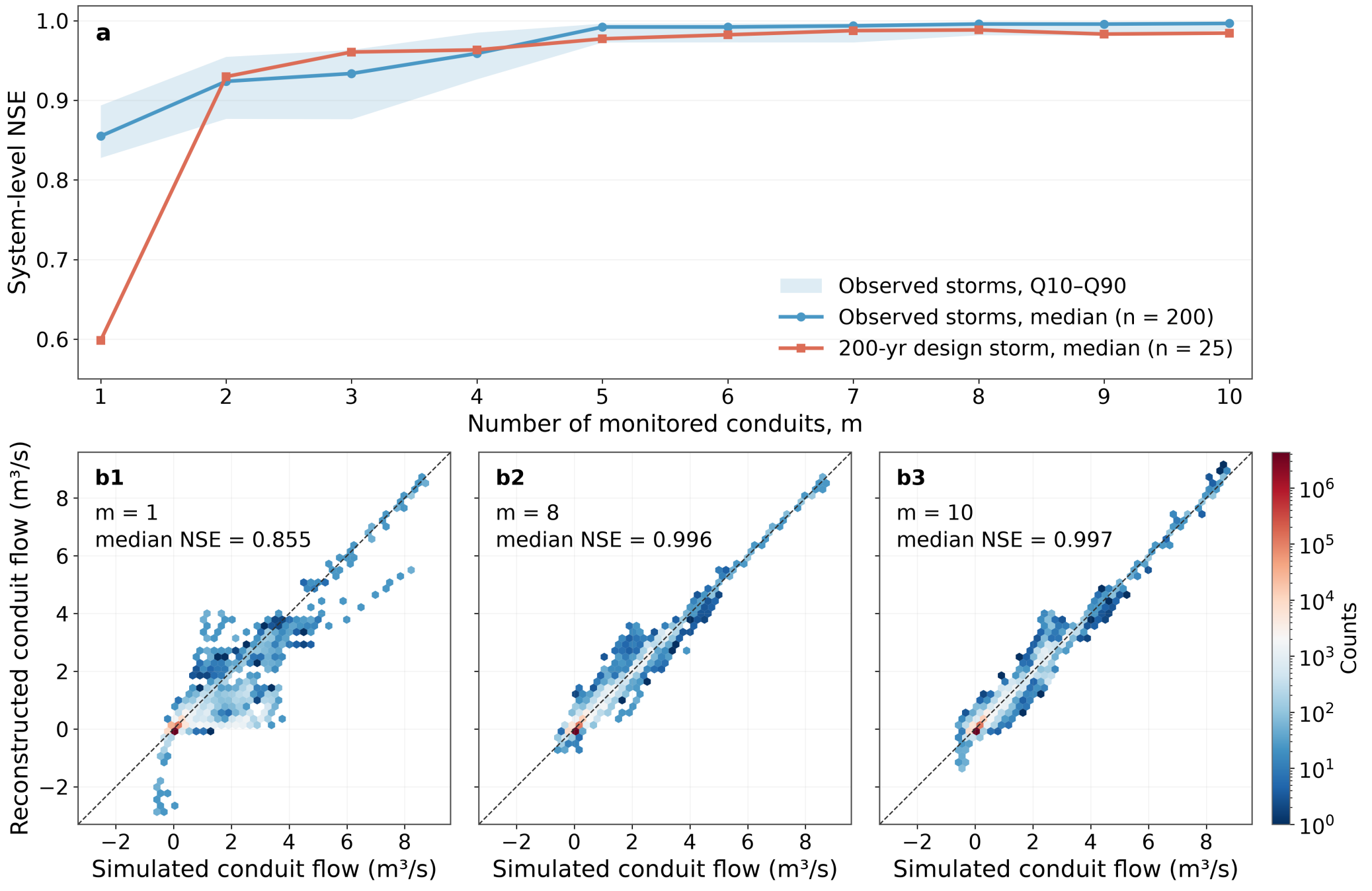

**Figure 4.** System-level reconstruction performance across the 225 held-out SWMM simulations. (a) Median and Q10–Q90 band of system-level NSE across 200 simulations from 8 observed rainfall events and median NSE across 25 simulations of the 200-year design event for number of monitored conduits $m$ = 1–10. (b1–b3) Simulated versus reconstructed conduit flow at (b1) $m$ = 1, (b2) $m$ = 8 and (b3) $m$ = 10, respectively, shown as hexagonal bin densities colored by the number of 5-min conduit flow values in each bin (log scale), pooled over the 76 conduits and all 225 held-out simulations (4.5 million values per panel); simulated flows above 2 $m^3/s$ occur only in the 200-year design storm. The dashed line is the 1:1 line, and the median NSE in each panel is that of the 200 observed-storm simulations.

More monitored conduits improved the representation of critical flow features, such as peak magnitude and flow volume. Figure 5 compares the DSS-reconstructed and SWMM-simulated flow trajectories and the relative errors of network-mean peak flow magnitude ($E_{Qp}$, %) and flow volume ($E_V$, %), together with the absolute peak-time difference ($|\Delta T_p|$, min), for $m$ = 1, 3 and 8 during the July 23 event (26.14 mm; highest 15-min intensity of the observed held-out events, 48.8 mm/h) and two other illustrated events (Table S4). Across the three events, as $m$ increased from 1 to 3 and 8, the error magnitudes decreased from 52.0–69.7% to 4.1–32.8% and 0.05–4.0% for $Q_p$, and from 32.6–39.1% to 7.4–14.6% and 0.45–1.4% for $V$. The absolute peak-time differences remained within one 5-min step at all three ranks.

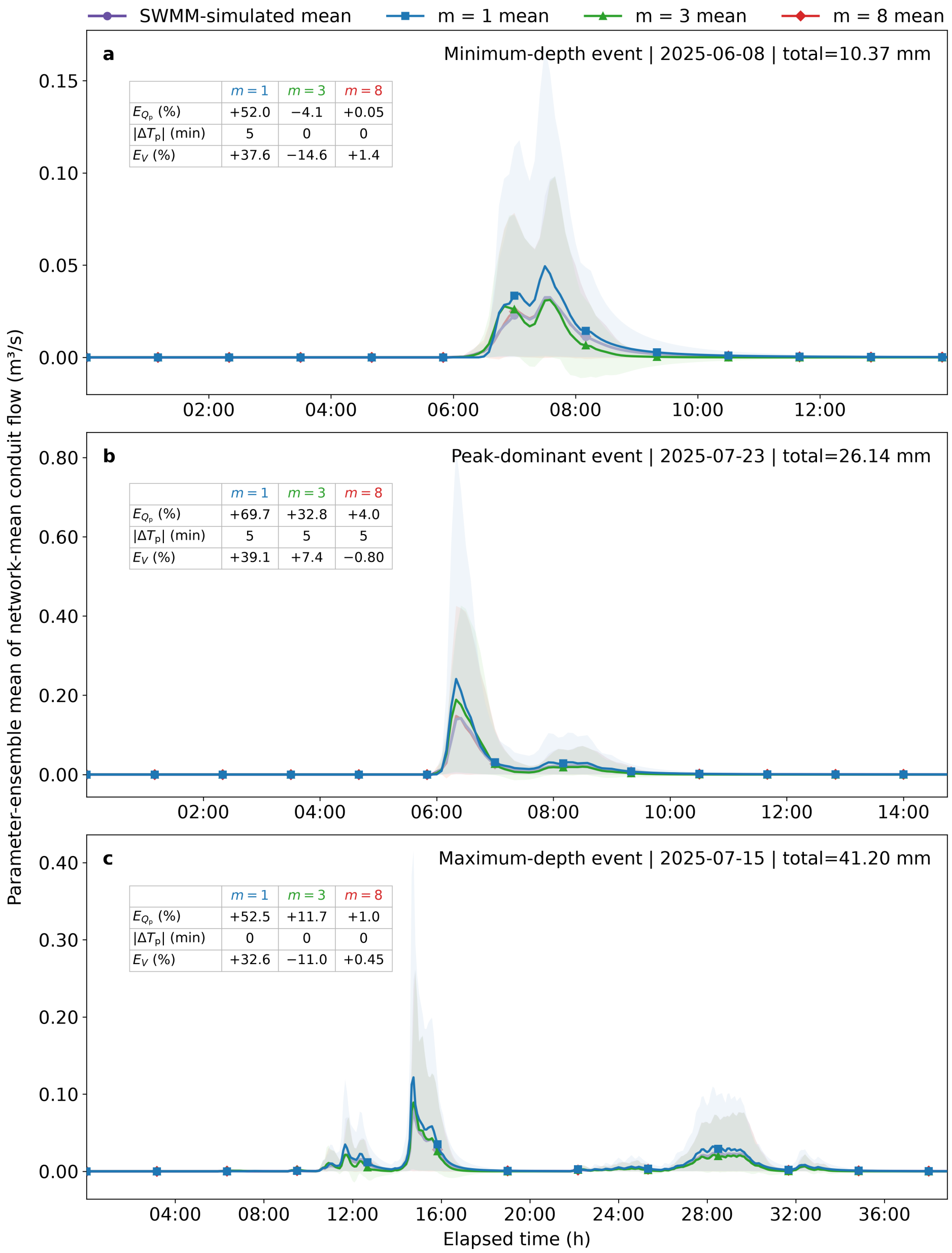

**Figure 5.** SWMM-simulated and DSS-reconstructed network-mean conduit-flow hydrographs for 3 observed held-out events: (a) June 8 (10.37 mm), (b) July 23 (26.14 mm), and (c) July 15 (41.20 mm), shown at number of monitored conduits $m = 1$, 3, and 8. At each conduit and time step, values were averaged across the 25 held-out plausible calibrated parameter sets; lines show the mean across all 76 conduits, and shaded bands show the pointwise 10th–90th-percentile range across conduits. Insets report relative peak flow error ($E_{Q_p}$, %), absolute peak-time difference ($|\Delta T_p|$, min), and relative flow volume error ($E_V$, %).

### 3.2. Conduit-Level Reconstruction Accuracy

Similar to system-level reconstruction, conduit-level reconstruction also generally improved as $m$ increased (Figure 6a; Table S6). As $m$ increased from 3 to 5, under the observed storms, the per-conduit median NSE of the active unmonitored conduits rose from 0.492 to 0.967 at the median and from −6.21 to 0.681 at the minimum. For $m \geq 5$, the conduit-level medians no longer increased monotonically because layouts are selected independently for each $m$, resulting in varying sets of unmonitored conduits. The medians were lower from $m = 5$ under the 200-year design storm (0.880 at $m = 8$; Figure 6a); this came from the 41–42 conduits active only under this storm, whereas the conduits also active under the observed storms kept medians of 0.97–0.99.

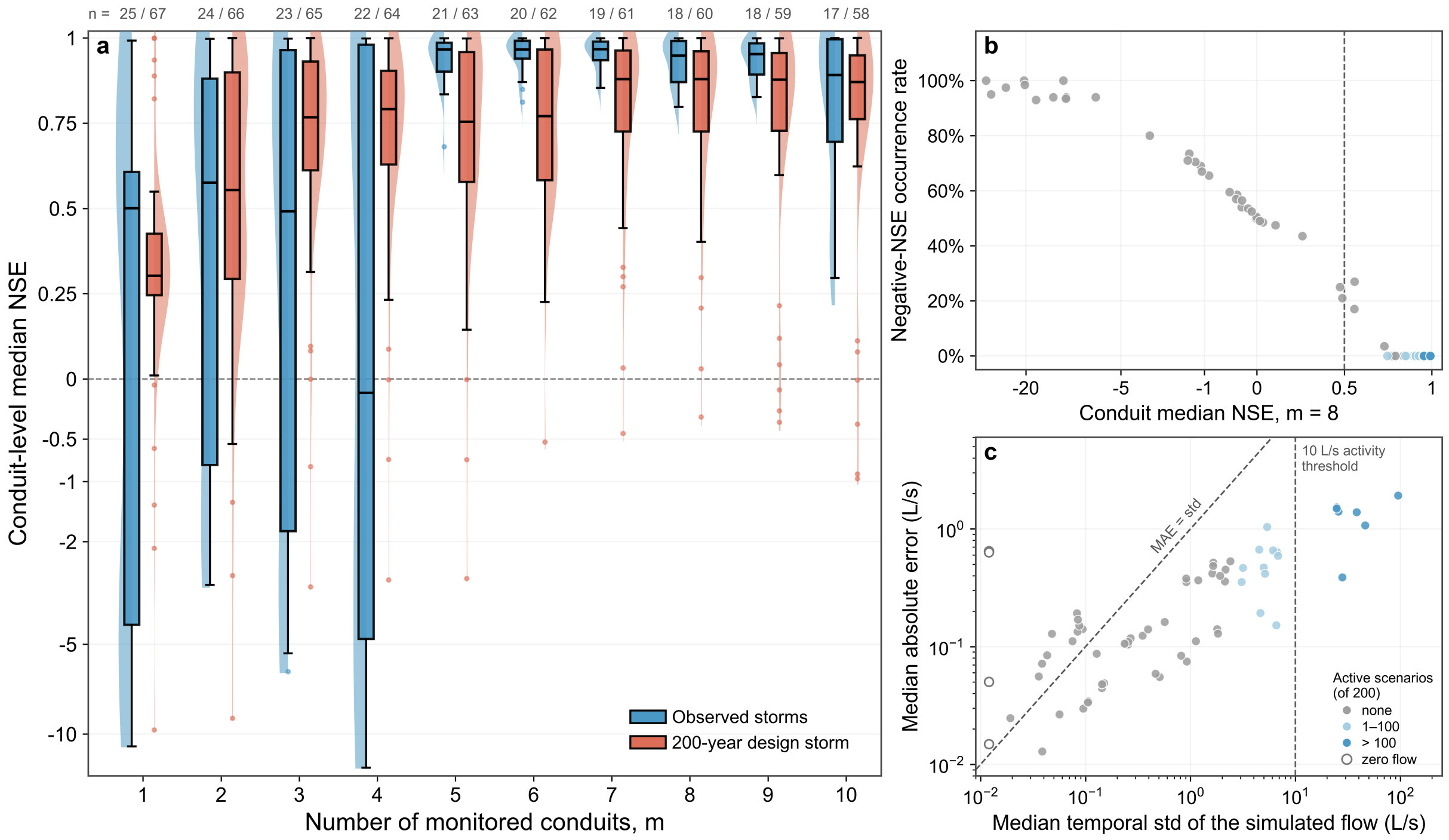

**Figure 6**. (a) Distributions of per-conduit median NSE of the unmonitored conduits, calculated from active cases within the 200 observed-storm (blue) and 25 design-storm (red) held-out simulations for $m$ = 1–10. Boxplots show medians and interquartile ranges, and adjacent half-violins show the corresponding distributions; the negative portions of the NSE axes in (a) and (b) use a signed-log scale. The numbers above the boxes give the conduits included for each storm type (observed / design). (b) Each unmonitored conduit's median NSE across all observed-storm simulations in which it is defined versus the percentage of those simulations with NSE < 0 at $m$ =

8. (c) Median absolute error versus median temporal standard deviation of the simulated flow for all 68 unmonitored conduits at $m$ = 8. A conduit–simulation pair was considered active when the simulated flow-rate temporal standard deviation was ≥ 10 L/s; colors in (b) and (c) give the number of simulations in which a conduit was active. The dashed line in (b) denotes NSE = 0.5; those in (c) denote absolute error equal to the temporal standard deviation and the 10 L/s activity threshold.

The reconstruction performance was higher at more "active" conduits. Using $m$ = 8 as an example, 18 of the 68 unmonitored conduits had at least one active observed-storm simulation, and the minimum of their per-conduit median NSE values was 0.798 over their active simulations (Figure 6a). However, 33 conduits that never became active under these storms had median NSE below 0.5 despite their smaller absolute errors (Figure 6b, c). The reconstruction performance's sensitivity to this threshold (from 0 to 20 L/s) can be found in Text S3 and Table S5.

### 3.3. Reconstruction Performance Across Sensor Placement Methods

QR-based layouts produced reconstruction performance comparable to GA-best across the tested number of monitored conduits ($m$), whereas Uniform Random placement performed substantially worse in most cases, with median NSE of 0.09–0.54 (Figure 7a). GA-best exceeded QR at $m$ = 1, 3, 4, 7 and 10, and QR exceeded GA-best at $m$ = 8; all paired differences were at most 0.016 (Figure 7c).

The largest-upstream-area heuristic failed with the DSS decoder (e.g., median NSE = −16.7 at $m$ = 3; Figure 7b). This is possibly because its first sites are consecutive conduits of the same trunk line, and the sampled basis was ill-conditioned (condition number 459 at $m$ = 10, versus 4.6 for QR; Figure S2b). With the event-weighted ridge decoder, the reconstruction performance of the largest-upstream-area heuristic improved markedly but still underperformed QR with the same decoder (e.g., paired ΔNSE = 0.113 at $m$ = 1 and 0.012 at $m$ = 8; Figure 7b, c). Although this gap was small at the system level, it was large at the conduit level: with this decoder and the area sites, 7–12 of the 16–21 active unmonitored conduits had median NSE below 0.5 for $m \geq 5$ (observed storms), versus none with the QR sites. On the QR sites, the event-weighted ridge decoder performed better than the DSS decoder at $m$ = 1–4 and the two decoders performed similarly for $m \geq 5$ (e.g., paired ΔNSE = −0.0009 at $m$ = 8).

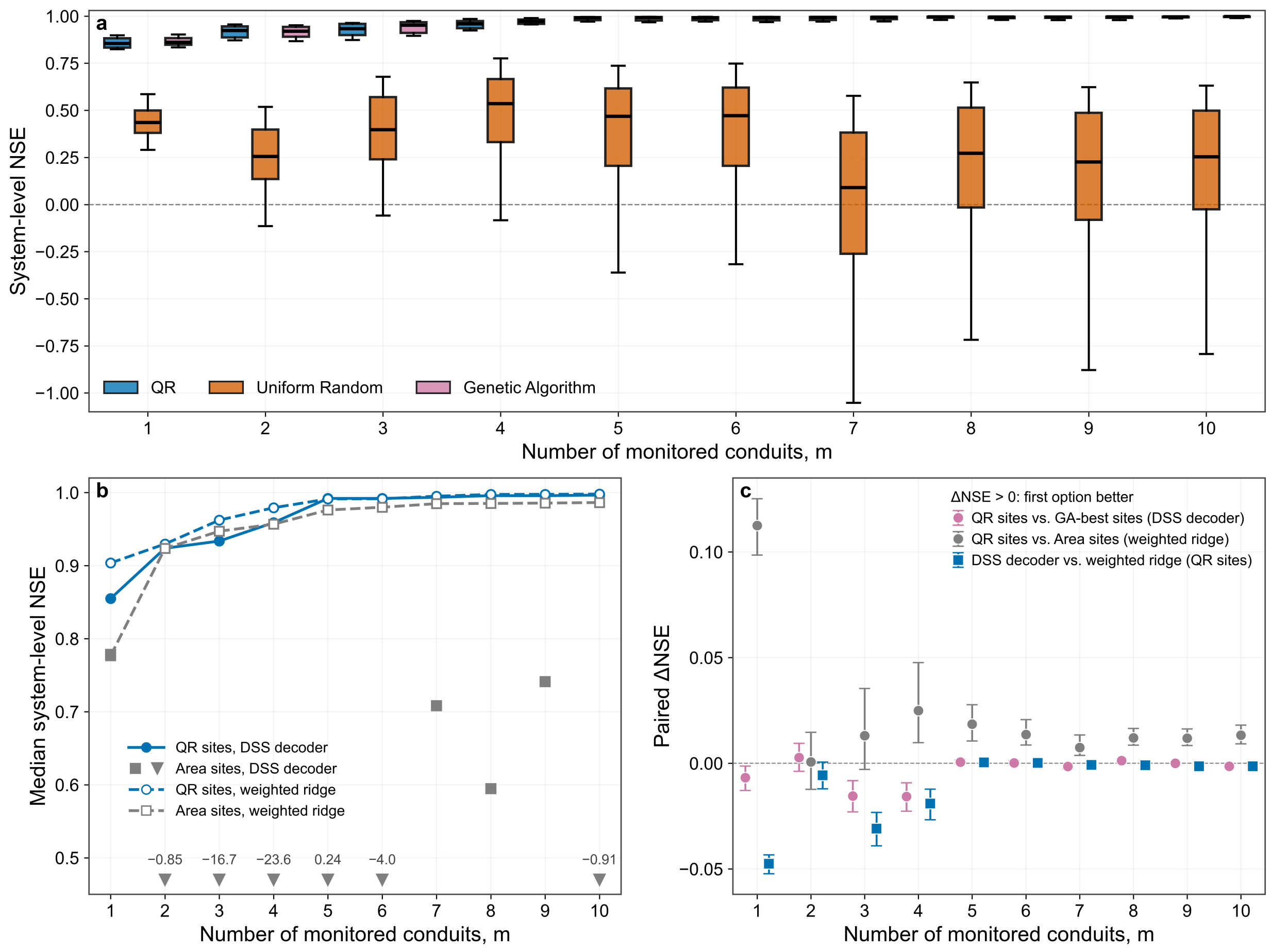

**Figure 7.** Comparison of held-out flow reconstruction performance, represented by system-level NSE, among QR and reference placement benchmarks. (a) QR, Uniform Random, and Genetic Algorithm (GA; best of 30 runs by the development objective) for $m$ = 1–10 on the 200 observed-storm simulations. The boxes show the median and interquartile ranges, and the whiskers show the 5th–95th percentiles; values beyond the whiskers are not shown. For Uniform Random placement, each held-out simulation is represented by the median NSE across the fixed candidate layouts (76 at $m$ = 1 and 2,048 at $m$ = 2–10). (b) Median system-level NSE versus m at the QR and largest-upstream-area (Area) sites, each with the DSS decoder and the event-weighted ridge decoder; medians below 0.5 at the Area sites with the DSS decoder are shown as triangles at the bottom of the axis with their values. (c) Paired differences in system-level NSE between the QR and GA-best sites (both with the DSS decoder), between the QR and Area sites (both with the event-weighted ridge decoder), and between the DSS decoder and the event-weighted ridge decoder (both at the QR sites); positive values mean that the first option performed better.

### 3.4. Effects of Measurement Errors on Reconstruction Accuracy

Random measurement noise had little effect on system-level reconstruction accuracy. Under multiplicative Gaussian noise, the median system-level NSE at $m = 8$ decreased from 0.996 under clean measurements to only 0.993 under ±15% three-sigma noise envelope (Figure 8a; full results in Figure S3a). Additive noise had a larger but still limited effect; at $m = 8$ it reduced the median NSE from 0.996 to 0.973 at 5 L/s and 0.916 at 10 L/s (Figure 8a; full results in Figure S3b), whereas the design-storm median remained 0.988 even at 10 L/s.

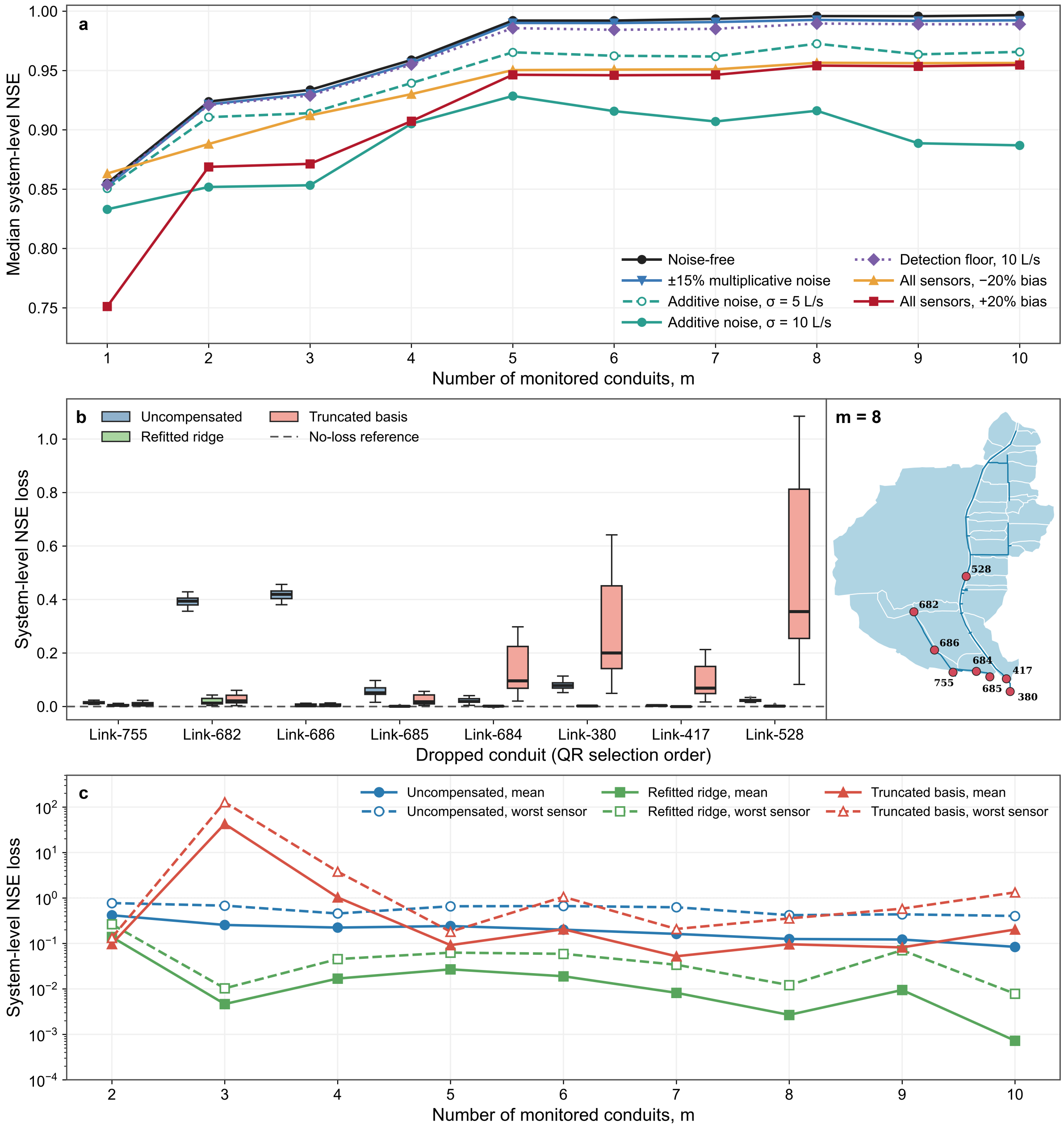

**Figure 8.** Flow reconstruction performance, represented by system-level NSE, under (a) representative measurement errors for $m$ = 1–10 (all cases in Figure S3), and reconstruction-performance change, represented by system-level NSE loss, after individually removing one monitored conduit from the rank-specific layouts at (b) $m$ = 8 (other $m$ in Figure S4), together with (c) the aggregated loss for $m$ = 2–10 (observed storms, n = 200). In panel (a), each line shows the median system-level NSE over the 200 observed-storm simulations (for random noise, each simulation value is the median over 1,000 realizations) under ±15% multiplicative Gaussian noise (three-sigma envelope), additive Gaussian noise with σ = 5 and 10 L/s, a 10 L/s detection floor, or a −20% or +20% bias on all sensors. In panel (b), the selected conduits are ordered according to the QR selection sequence, each with the losses under the uncompensated, refitted-ridge and truncated-basis cases, and the dashed line denotes the no-loss reference. The map on the right illustrates their network locations. Panel (c) summarizes the mean and worst-case per-sensor median loss after removing one monitored conduit at each $m$ for the uncompensated, refitted-ridge and truncated-basis cases.

Of the systematic errors, a 10 L/s detection floor had little effect, while a −20% or +20% bias on all sensors had a larger effect because the linear DSS decoder then scales every reconstructed flow by the same 20%. At $m$ = 8, it lowered the median NSE by approximately 0.04, to 0.957 (−20% bias) and 0.954 (+20% bias) (Figure 8a). The same bias on any single sensor mattered less (median NSE ≥ 0.978; Figure S3c).

### 3.5. Effect of Monitored Conduit Loss on Reconstruction Performance

The flow reconstruction performance showed some sensitivity to measurement loss, but the effect was strongly location dependent (Figure 8b; other $m$ in Figure S4). Using $m$ = 8 as an example, the individual removal of 6 of the 8 monitored conduits resulted in median NSE losses below 0.08, whereas the removal of Link-686 or Link-682 produced substantially larger losses of 0.419 and 0.393. The refitted ridge decoder reduced the mean and worst-case losses at $m$ = 8 to 0.003 and 0.012.

Figure 8c further illustrates the mean and worst-case reductions in reconstruction performance for $m$ = 2–10. The mean uncompensated loss generally decreased with $m$, from 0.416 at $m$ = 2 to 0.125 and 0.084 at $m$ = 8 and 10. In contrast, the worst-case performance loss did not show obvious reduction and remained substantial at high $m$: 0.772, 0.419, and 0.402 at $m$ = 2, 8 and 10, respectively.

Truncating the basis reduced the loss at some $m$ but increased it at others. The refitted ridge decoder reduced the loss across all tested $m$, particularly for $m \geq 3$; it kept the worst-case loss at or below 0.071 for $m \geq 3$ (Figure 8c).

# 4. Discussion

## 4.1. Sensor Placement Optimization in Urban Drainage Networks

Prior studies have shown that optimized sensor placement can substantially reduce monitoring requirements, although the required number of sensors depends on the monitoring objective and network characteristics. Grimaldi et al. (2024) showed that information from 4 selected sub-basins could achieve discharge-forecasting performance comparable to that from 26 potential monitoring cross-sections. Huang et al. (2025) jointly optimized sensor placement and model calibration for a 54-node drainage network using five sensors, and Wang et al. (2023) attained maximal coverage with 28–38 sensors in a 786-node network. Recently, Zheng et al. (2025) combined information-theoretic sensor placement with matrix completion of the water-level matrix and, applied to a drainage network with 878 nodes, achieved flooding risk assessment accuracy of 76% and 82% using 4 and 8 sensors, respectively.

In the present network of 76 conduits, DSS achieved a median system-level NSE of 0.992 with only 5 monitored conduits across the observed-storm simulations. Thus, about 5 conduits sufficed to reconstruct network-wide flow, while also remaining robust to moderate measurement noise. The placement benchmark further supported pivoted QR as a practical placement strategy within DSS because it constructs rank-specific layouts directly and avoids the redundant trunk-line sites of the area heuristic, whereas genetic-algorithm methods require iterative candidate-layout optimization. Another key contribution of DSS lies in its low-dimensional basis – where five modes held 99.2% of the event-weighted development energy – and its one-step site selection, with a decoder that can be regularized when sensors are few (Erichson et al., 2020; Karnik et al., 2026) or when one fails.

Direct comparison of required sensor counts between this study and other existing studies is difficult because they all have different study objectives, including flood-awareness coverage, water-level prediction, parameter calibration, and sensor-network evaluation. DSS can complement traditional localized or discrete point-based monitoring and may suit flashy urban drainage systems, where reconstructing continuous hydraulic transients and peak timing is critical. In practice, a utility with a calibrated hydraulic model can apply DSS before installing sensors, and the sensor-loss results indicate which sites warrant priority maintenance or redundancy.

### 4.2. Sparse Reconstruction of Urban Drainage Flows Under Held-out Conditions

A DSS basis and sparse layout learned once from the reference calibrated SWMM simulations retained strong reconstruction skill when applied unchanged to modeled drainage states generated under held-out hydrologic forcings and conditions. Across the 200 observed-storm simulations, the $m = 8$ configuration achieved a median system-level NSE of 0.996 and a 10th-percentile NSE of 0.982. These results indicate that the dominant spatial flow patterns identified from the 17 development storms remained informative across the tested rainfall and parameter conditions.

The strong DSS-based flow reconstruction is consistent with substantial coherence in the modeled drainage response. Rainfall provides a common forcing across the catchment, while runoff generation and routing through the connected storm-sewer network generate hydraulic correlations among conduits (He et al., 2025a; Zhang et al., 2024), so that much of the modeled response was represented by a relatively low-dimensional spatial basis (Text S6). Event weighting prevented the design storms from dominating the basis and improved the cross-validated reconstruction accuracy.

### 4.3. Limitations and Future Work

The DSS-based reconstruction was developed and evaluated entirely using SWMM-simulated conduit flows rather than directly observed internal drainage states. The reported reconstruction accuracy should therefore be interpreted as recovery of modeled network states, and direct field validation remains necessary.

More work is needed to apply DSS across multiple networks with different hydraulic conditions and more spatially variable rainfall forcings. Future work should evaluate DSS in drainage systems with long-term, spatially distributed monitoring. Such observations could be used to construct the SVD basis, select monitoring locations, and directly test sensitivity to measurement errors and data loss.

## 5. Conclusion

Accurate monitoring of distributed sewer hydraulic states under limited resources is essential for real-time operation and mitigation. This study integrated process-based simulation via EPA-SWMM with data-driven sparse sensing (DSS) to optimize sensor placement for sewer flow reconstruction. Our results demonstrate that DSS can exploit known patterns in simulated flow data to identify appropriate locations for flow monitoring and use these sparse data to reconstruct flow rate conditions across the entire network.

With 5 monitored conduits considered for reconstruction (equivalent to 6.6% of the network), the flow rates across the network were reconstructed with a system-level median NSE of 0.992 and a $10^{th}$ percentile of 0.973 for observed storms (median 0.967 at the active unmonitored conduits). Pivoted-QR-selected layouts were competitive with genetic-algorithm layouts and better than largest-upstream-area placement. The DSS flow reconstruction performance showed little sensitivity to prescribed multiplicative Gaussian measurement noise, additive noise up to about 2 L/s and a 10 L/s detection floor. It showed location-dependent responses to loss of a monitored conduit, which, however, can be largely mitigated via a refitted regression decoder.

Taken together, this simulation-based evaluation supports the use of DSS for designing sparse flow-monitoring networks where a calibrated hydraulic model is available. Moving forward, to test its broader applicability, DSS should be tested with field observations of conduit flow, as well as across multiple networks with different hydraulic conditions and more spatially variable rainfall regimes.

## Acknowledgements

The authors appreciate Tim Olson from Bolton & Menk Inc. for sharing their EPA-SWMM model. Kun Zhang also acknowledges the Minnesota Water Resources Center’s Watershed Innovation (WINS) Grant Program and the Swenson College of Science and Engineering’s Dr. Howard Higholt professorship for supporting his research at UMD. Regents of the University of Minnesota gratefully acknowledges that The Water Research Foundation are funders of certain technical information upon which this publication is based. Regents of the University of Minnesota thanks The Water Research Foundation for their financial, technical, and administrative assistance in funding the project through which this information was discovered. This material does not necessarily reflect the views and policies of the funders, and any mention of trade names or commercial products does not constitute the funders' endorsement or recommendations thereof.

## Data Availability Statement

Data and scripts required to reproduce the reported analyses are available at the GitHub repository (https://github.com/DzhZzz/Data-Driven-Sparse-Sensing); the proprietary SWMM model is not publicly distributed.

## Funding

This project is directly supported by the Watershed Innovation (WINS) Grant Program #00120120 ("Urban Flood Prediction from Limited Data: A Data-Model Integrated Approach") through USGS Water Resources Research Act and The Water Research Foundation 5304 program ("Optimizing Nature-based Solutions at the Watershed Scale with Real-time Sensing and Control").

## CRediT authorship contribution statement

Zihang Ding: Conceptualization; Data curation; Formal analysis; Investigation; Methodology; Validation; Visualization; Writing – original draft; Writing – review & editing. Amit Kumar: Validation, Writing – review & editing. Imran Md. Azizul Islam: Validation, Writing – review & editing. Mila Avellar Montezuma: Writing – review & editing. Ruihang Zhang: Validation, Writing – review & editing. Kun Zhang: Conceptualization; Funding acquisition; Methodology; Project administration; Resources; Software; Supervision; Validation; Writing – review & editing.